\documentclass[conference]{IEEEtran}
\IEEEoverridecommandlockouts
\usepackage{cite}
\pdfoutput=1 
\usepackage{amsmath,amssymb,amsfonts}
\usepackage{algorithmic}
\usepackage{graphicx}
\usepackage{textcomp}
\usepackage{xcolor}
\usepackage{caption}
\usepackage{tikz}

\newcommand{\chronogram}[4]{
  \draw[thick] (0, #1) node[left]{$V_{#1}(t)$}
     -- ++(#2,  0.0) -- ++(0,  +0.5) -- ++(#3,  0.0) -- ++(0,  -0.5) -- (#4,  #1);
   \draw[thin,<->] (#2, #1+0.25) -- node[below,,midway]{$d_{#1}$} (#2+#3, #1+0.25);
   \draw[thin,dotted] (#2, #1) -- (#2, 0);
   \draw[thin] (#2, 0) -- (#2, -0.1) node[below]{$t_{#1}^{+}$};
   \draw[thin,dotted] (#2+#3, #1) -- (#2+#3, 0);
   \draw[thin] (#2+#3, 0) -- (#2+#3, -0.1) node[below]{$t_{#1}^{-}$};
}

\usepackage{url}
\usepackage{parskip}

\def\BibTeX{{\rm B\kern-.05em{\sc i\kern-.025em b}\kern-.08em
    T\kern-.1667em\lower.7ex\hbox{E}\kern-.125emX}}
\begin{document}

\title{Exploring the role of structure in\\ a time constrained decision task}

\makeatletter
\newcommand{\linebreakand}{%
  \end{@IEEEauthorhalign}
  \hfill\mbox{}\par
  \mbox{}\hfill\begin{@IEEEauthorhalign}
}
\makeatother

\author{
\IEEEauthorblockN{Naomi Chaix-Eichel}
\IEEEauthorblockA{
\textit{Inria Bordeaux Sud-Ouest}\\
\textit{Univ. Bordeaux, CNRS } \\
}
\and
\IEEEauthorblockN{Gautham Venugopal}
\IEEEauthorblockA{
\textit{International Institute of }\\
\textit{Information Technology, Hyderabad} \\
}
\and
\IEEEauthorblockN{Boraud Thomas}
\IEEEauthorblockA{
\textit{IMN, Bordeaux}\\
\textit{Univ. Bordeaux, CNRS } \\
}
\and

\IEEEauthorblockN{Nicolas P. Rougier}
\IEEEauthorblockA{
\textit{Inria Bordeaux Sud-Ouest}\\
\textit{Univ. Bordeaux, CNRS } \\
}
}

\maketitle

\begin{abstract}
The structure of the basal ganglia is remarkably similar across a number of species (often described in terms of direct, indirect and hyperdirect pathways) and is deeply involved in decision making and action selection. In this article, we are interested in exploring the role of structure when solving a decision task while avoiding to make any strong assumption regarding the actual structure. To do so, we exploit the echo state network paradigm that allows to solve complex task based on a random architecture. Considering a temporal decision task, the question is whether a specific structure allows for better performance and if so, whether this structure shares some similarity with the basal ganglia. Our results highlight the advantage of having a slow (direct) and a fast (hyperdirect) pathway that allows to deal with late information during a decision making task.\end{abstract}


\section{Introduction}
The structure of the Basal Ganglia (BG) is remarkably similar across a number of species, from the newt to the primate\cite{Boraud:2018}. These ganglia are often described in terms of pathways, including the direct, indirect and hyperdirect pathways. The role of each pathway is still under scrutiny and consequently, there exist several hypothesis regarding their role in action selection and decision making for which basal ganglia are known to be deeply involved. In this article, we are primarily interested in exploring the role of structure when solving a decision task while avoiding to make any strong assumption regarding the actual structure. To do so, we exploit the echo state network (ESN) paradigm that allows to solve complex tasks based on a random architecture\cite{jaeger2007echo}. Considering a temporal decision task, the question is whether a specific structure allows for better performance and if so, whether this structure shares some similarity with the basal ganglia. Unfortunately, we cannot explore each and every variant of architecture because the number of different structures for a fixed number $n$ of neurons is huge (and grows exponentially with $n$). Instead, we restrict our exploration to a much smaller subset where a model is built around two pathways, each of them being made of several chained ESN and in charge of processing a single input. We also added a continuous case based on topological reservoir that allows to have distance based connectivity patterns and allows us to take the limit of the two pathways structure. These models are loosely inspired from the direct and hyperdirect pathways of the BG\cite{schmidt2013canceling}, with the latter allowing the production of a fast ”stop signal” thanks to a reactive inhibition.

\subsection{Structured ESN}

The role of structure in ESN have already been addressed in a number of works. While \cite{dale2021reservoir} quantified how structure affects the behavioral characteristics of the ESN, several studies have demonstrated that replacing the initial random topology of the ESN by more organized structures could improve the overall performances of the model. Nonetheless, rather than completely removing the randomness of the network topology, certain structures allows to combine both random and structured connections. One well-known example is the small world network, which has been observed in the neural network of the C. Elegans\cite{watts1998collective} and in other brain systems\cite{bassett2006small}.  \cite{cheng2015efficient,bai2017reservoir} have shown that incorporating small-world structure into ESNs results in performance improvements on benchmark tasks. Various other structures for ESN  have also demonstrated significantly superior performance, including the combination of scale-free and small-world networks\cite{deng2007collective,kawai2019small,kitayama2022guiding}. Additionally, modular structures \cite{rodriguez2019optimal},  forward topology with shortcuts pathways \cite{dominey2022effects} and hierarchically clustered ESNs \cite{jarvis2010extending} have been explored, each impacting memory capacity, temporal properties, and reservoir stability. An alternative approach known as Deep Reservoir Computing, involves investigating various structures though the combination of multiple random reservoirs rather than a single one \cite{gallicchio2017deep,moon2021hierarchical}, \cite{xue2007decoupled}. Our study aims to contribute to the existing findings by incorporating the application of Deep Reservoir Computing and the utilization of ESN with a forward topology.

\subsection{Time-constrained decision making}
The temporal aspect of the decision making task used in this work is a significant component in making the task challenging to solve and interesting to study. In the real world, decision making is time constrained. Decisions need to be taken within certain timeframes, where the importance of speed and the need for caution can vary across situations. In many such cases, there would exist a speed-accuracy tradeoff, where one can collect more information or ponder more over the choice in order to make a better decision at the cost of time. As navigating such trade-offs optimally would be important for one's survival \cite{foragingESN}, many sophisticated models have been developed to model animal behavior in such situations. A popular set of models used to study how animals approach time-constrained decisions are Evidence Accumulator Models (EAMs)\cite{Ratcliff2008,Brown2008}. In such models, deliberating over a decision is modeled as accumulating observations which over time can be perceived as evidence for making one or another choice. When the accumulated evidence reaches a certain threshold, a decision is taken. 
Such models are able to seamlessly integrate the myriads of factors that affect animal decisions, with the threshold indicating response caution, and observations serving as probabilistic likelihoods for choices. Electrophysiological evidence for correlated ramping signals in specific brain regions has been observed \cite{Pisauro2017}. Another widely used EAMs are the Drift Diffusion Mode, which is equivalent to the Wald Sequential Probability Ratio Test. Despite the many factors in favor of EAMs, one occasion where such models fail is when new evidence emerges, necessitating quick decision changes \cite{Cisek2009}. EAMs struggle in promptly adjusting to sudden changes, especially when the new evidence contradicts previously acquired information. While alternative models such as Leaky Accumulator Models \cite{Usher2001} and Urgency Gating Models \cite{Thura2012} have been proposed as a solution to this problem, they often don't provide as good fits to animal behaviour when considered across a wide variety of tasks.  This paper introduces an alternative approach to address this challenge, with the objective of building ESNs with multiple architectures, each capable of handling temporal information in a distinct manner.

\section{Methods}
\label{sec:methods}
\subsection{Tasks}

We consider a non-stochastic two-arm bandit task where an agent is presented with two options, each associated with a certain amount of reward. The options are presented for a fixed duraction (30 timesteps) after which, the agent receives a reward based on the amount attached to its choice. This trivial task is further complexified by introducing a choice indirection (motor aspect) and differential timing (temporal aspect). \\


\subsubsection{Motor aspect} An option is represented by a stimulus with a given identity (1 to 4) and a given position (1 to 4). For a given trial, stimuli identity and position are mutually exclusive. The value of an option is solely attached to the identity of the stimulus, irrespective of its position. The agent's choice is interpreted as a position from which the identity of the simulus can be retrieved (and hence the amount of reward).\\


\subsubsection{Temporal aspect} The onset and offset times of the two options are independent. This means that they may or may not have the same duration, they may or may not start nor finish at the same time and they can be completely disjoint, i.e. there is no overlap between the two options. This temporal aspect considerably complexifies the task. In some trials, the agent must maintain the value of the first option (working memory) while in other trials, the agent has to deal with a late but better option (time constrained decision). In this study, we only consider ovelapping stimuli.


\begin{figure}[htbp]
  \includegraphics[width=\columnwidth]{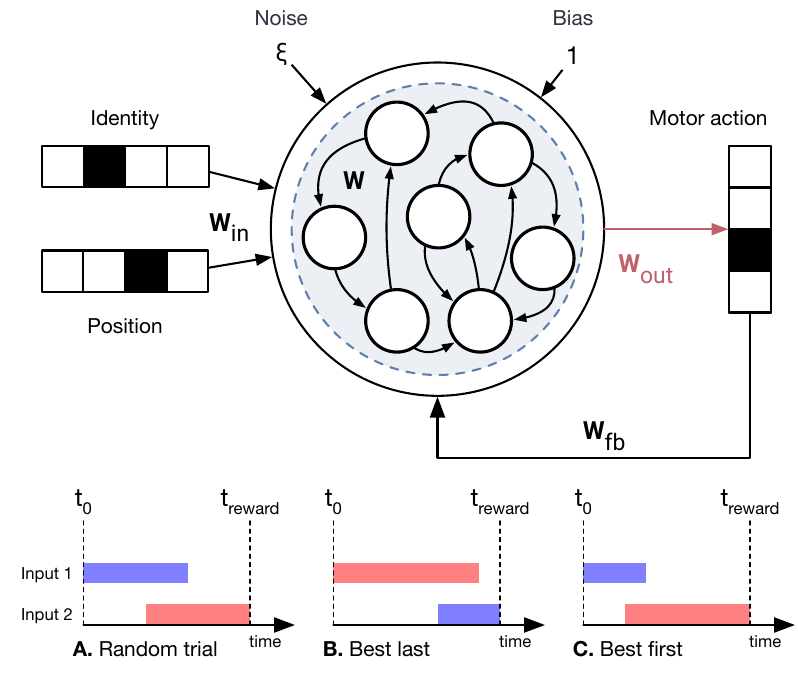}
  \caption{\textbf{Top} Model architecture with a motor output (direction of movement). The black arrows are fixed and the red is plastic. \textbf{Bottom} The red (worst) and blue (best) stimuli can have different onset/offset times and the reward is receives at a fixed time. }
  \label{fig:ESN}
\end{figure}

\begin{figure}[htbp]
  \includegraphics[width=\columnwidth]{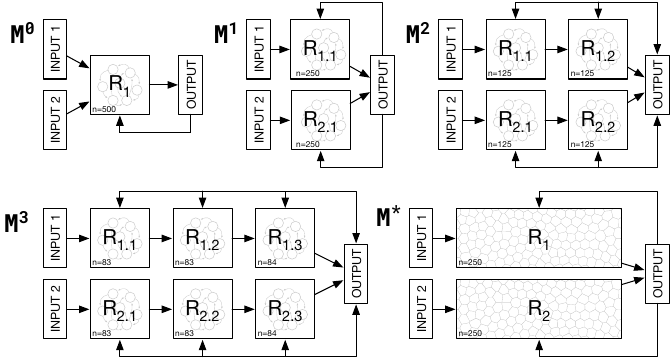}
  \caption{Model are composed of one to several chained ESN, all connected to the readout, all receiving feedback from the output. {\textbf M$^\mathbf{0}$}: Regular ESN. {\textbf M$^\mathbf{1}$}: Dual pathway each made of a single ESN. {\textbf M$^\mathbf{2}$}: Dual pathway each made of two chained ESNs. {\textbf M$^\mathbf{3}$}: Dual pathway each made of three chained ESNs. {\textbf M$^\mathbf{*}$}: Dual pathway each made of one continuous ESN.}
  \label{fig:models}
\end{figure}

\subsection{Models}


An Echo State Network (ESN) is a specific type of reservoir computing\cite{jaeger2007echo} that is a recurrent neural network composed of randomly connected units, associated with an input and an output layer. Only the output neurons, referred as the readout neurons are trained, as depicted with the red arrow in Figure \ref{fig:ESN}. The neurons have the following dynamics:
\begin{equation}
\frac{1}{\alpha} \frac{d \bf x}{dt} = -{\bf x} + \tanh(W.{\bf x} + W_{in}.{\bf u} + W_{fb}.{\bf y}) 
\end{equation}
\begin{equation}
 {\bf y} = W_{out}.\bf x
\end{equation}
where \({\bf x}\), \({\bf u}\) and \({\bf y}\) represent the reservoir states, input, and output. \(W\),  \(W_{in}\), and  \(W_{out}\),\(W_{fb}\) are weight matrices, while \(tanh\) refers to the hyperbolic tangent function. $\alpha$ refers to the leak rate, a crucial parameter of the ESN that plays a role in controlling the memory and timescale of the network's dynamics: a small leak rate indicates a bigger memory and a slower dynamics, whereas a big leak rates lead to a smaller memory but a higher speed of update dynamics\cite{lukovsevivcius2012practical}. All models have been implemented using the Python library ReservoirPy\cite{trouvain2020reservoirpy}.\\


\subsubsection{Architecture} From the classical ESN (Figure \ref{fig:ESN}), we derived several architectures (Figure \ref{fig:models}) that are all characterized by the presence of two distinct pathways, a slow pathway and a fast pathway, drawing inspiration from the direct and hyperdirect pathways of the basal ganglia \cite{schmidt2013canceling}, with the latter allowing the production of a fast "stop signal" thanks to a reactive inhibition. This "stop emergency brake"\cite{aron2006cortical} is attributed to the significant role of the nucleus STN. Each of these two pathways receive segregated inputs, that is, each pathway receives a single option. More precisely, the slow pathway receives the earliest option and the fast pathway receives the latest option. All the models possess a total of 500 neurons equally distributed across multiple reservoirs, with all reservoirs connected to the readout and receive feedback signals from it.


We also designed a model (M$^\mathbf{*}$) that is equipped with a topology \cite{Rougier2018} such that it is possible to constrain activity propagation along a feed-forward axis (from input to output). This allows the reservoir to progressively process information along the main axis, where early units (that are closer to the input) have access to local and recent information while late units have access to global and processed information. The output layer which has access to both early and late units has the ability to accumulate information and take accurate decisions, while at the same time having the ability to quickly respond to changes in the environment. To make these type of reservoir, the distribution neurons across a 2D space is first defined by using the algorithm described in \cite{Rougier2018} from which the connectivity matrix can be derived. Individual connections are established based on  the nearest neurons that meet angle constraints as shown in figure \ref{fig:topology}, connections are established between input and output neurons following a rule in which the probability of connection exponentially decreases with distance.

\begin{figure}[htbp]
 \centering
  \includegraphics[width=\columnwidth]{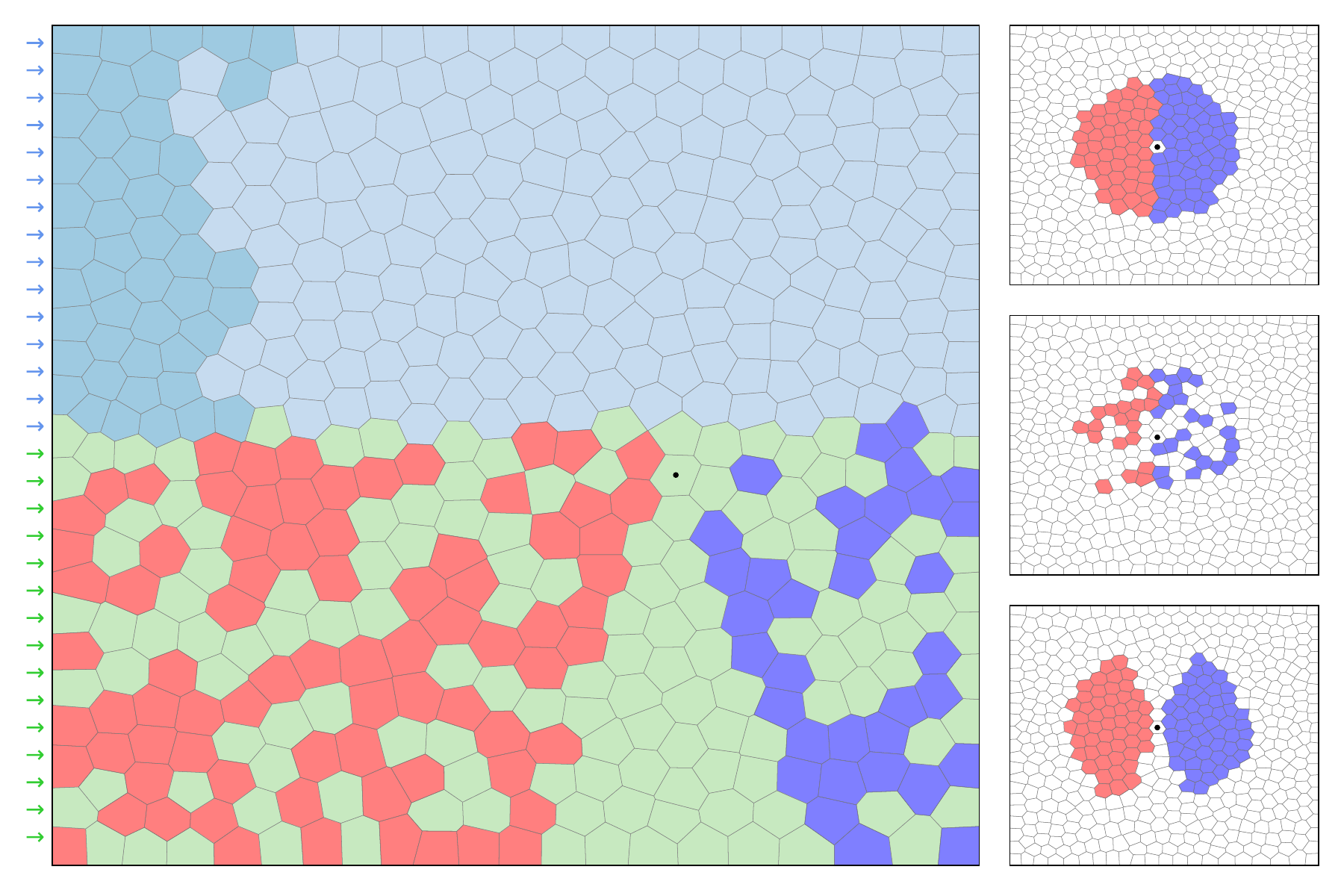}
  \caption{
  \textbf{Left} Internal structure of the M$^\mathbf{*}$ model. The two pathways are shaded in blue (top) and green (bottom) and are completely segregated (no reciprocal connections between them). The input to the top pathway has ben shaded in darker blue while the output is not represented for clarity because output receives connection from virtually all units. The input to the bottom pathway is similar but not represented for clarity. Instead, a typical unit (dot) connection pattern is represented with red for incoming connections and blue for outgoing projections.
  \textbf{Right} The connection pattern of a unit is governed by an angle $\theta$ (ranging from 0 to 90), a fixed radius $r$ and a connection probability $P_c$. Top) $\theta=90^{\circ}$, $P_c=1.0$ Middle) $\theta=90^{\circ}$, $P_c=0.4$ Bottom) $\theta=70^{\circ}$, $P_c=1.0$}
\label{fig:topology}
\end{figure}
\subsubsection{Learning} The readout layer is trained using online reinforcement learning (RL) based on equation \ref{eq:train_1} and \ref{eq:train_2}, where only the weights associated with the selected choice undergo updates. The choice of RL as the learning rule comes from its biological plausibility, given that cortico-basal-ganglia (BG) circuits are trained through reinforcement, thanks to the encoding of reward prediction error (RPE) with dopamine\cite{bar2003information}. Equations read:
\begin{equation}
W_{out}(choice) = W_{out}(choice) + \delta W_{out}
\label{eq:train_1}
\end{equation}
\begin{equation}
\delta W_{out} = \eta.(r-{softmax({\bf y}, \beta)}[choice]).({\bf x}-{\bf x_{th}})
\label{eq:train_2}
\end{equation}
%
%
 where $choice$ represents the index associated with the model's chosen action. $\eta$ is the learning rate. The function $softmax(y, \beta)$ applies softmax to the model's output with $y$ as the model's output and $\beta$ as a parameter. $x_{th}$ denotes a small constant value, and $r$ corresponds to reward feedback received. Each stimulus identity is associated with a specific fixed reward value, which can be 1, 0.75, 0.5, or 0.25. The action selection process follows the epsilon-greedy method, allowing to balance between exploitation and exploration phases. When the agent is in the exploitation mode, it selects the action that corresponds to the highest output value of the model ($argmax(output_{model})$). In contrast, during exploration, the agent randomly selects one action from the set of all available actions, with equal probability among the four possible choices. The method uses a parameter called epsilon ($\epsilon$), which starts at 1 during the beginning of each simulation and ends at 0, signaling a shift towards exclusive exploitation of learned knowledge. This dynamic $\epsilon$ adjustment enables the agent to transition from exploration to exploitation.\\


\subsubsection{Optimization} All models undergo a hyperparameter optimization process using the Optuna library \cite{akiba2019optuna}. More specifically, spectral radius ($sr$), leak rate ($\alpha$), input connectivity ($W_{in}$),  output sparsity $W_{out}$, the reservoir connectivity $W$, exploration rate  ($\beta$)  and the learning rate ($\eta$) are optimized. For the \textbf{M}$^\mathbf{*}$ model, rather than the spectral radius and reservoir connectivity, the connections are determined by the radius, angle and sparsity parameters which are optimized instead. The spectral radius parameters is only applicable when the connection angles are set to a value greater than 90$^\circ$, as when they are lesser than 90$^\circ$, there are no recurrent connections in the reservoir, and thus no possibility of chaotic activity as any input would inevitably decay. In the case of angles greater than 90$^\circ$, we found that constraining the weights based on spectral radius had a detrimental effect on the performance of the network. We believe that due to the input being presented from one side of the network and the unique nature of connectivity in the reservoir, not all eigenvectors of the reservoir connectivity matrix may become instantiated in the network. Thus, as scaling the weights based on the largest eigenvalues were not giving the best results, the weights of the topological reservoirs have not been scaled using a spectral radius in this work. The optimization process consists in running 600 simulations with different set of parameters sampled using Tree-structured Parzen Estimator (TPE) \cite{bergstra2011algorithms}. Each simulation consists of 1000 trials, and performance assessment occurs over the last 200 trials of the simulation by counting the number of successful actions (best option chosen). 

\subsection{Protocol}

\begin{figure}[htbp]
\begin{center}
\begin{tikzpicture}
  \def\t0{0}
  \def\treward{6.5}
  \draw[-latex] (0, 0) -- ++(\treward+1, 0) node[right]{$t$};
  \draw[] (0, 0) -- ++(0, 3);
  \foreach \i in {0,0.1, ..., 7.0} \draw[thin] (\i,0)--(\i,.05);
  \chronogram{1}{1.0}{2}{\treward}
  \chronogram{2}{3.5}{2}{\treward}
  \draw[thin,dotted] (3.0, 1.5) -- (3.0, 3.5);
  \draw[thin,dotted] (3.5, 1.5) -- (3.5, 3.5);
  \draw[thin,dotted] (\treward, 3) -- (\treward, 0);
  \draw[thin] (\t0, 0) -- (\t0, -0.1) node[below]{$t_{0}$};
  \draw[thin] (\treward, 0) -- (\treward, -0.1) node[below]{$t_{reward}$}; 
\end{tikzpicture}
\end{center}
\caption{The two stimuli $V_i$ are characterized by their respective onset ($t_i^+$) and offset ($t_i^-$) time. The time of decision $t_{reward}$ is fixed and constant across trials.}
\label{fig:chronogram}
\end{figure}
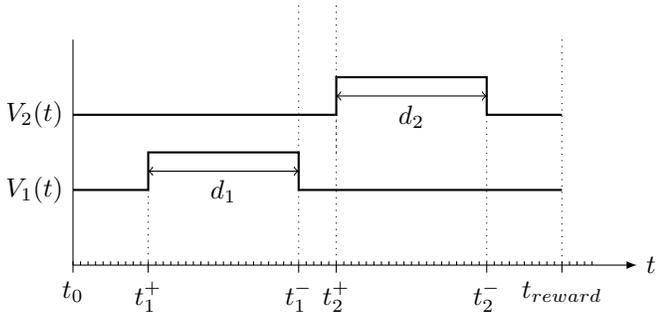
The models were optimized across a broad spectrum of timing and delay conditions (see Figure \ref{fig:chronogram}). While $t_{t_1^+}$ and $t_{reward}$ are fixed and respectively set to 5 and 1, $d_1$ and $d_2$ vary within the range of 5 to 20, and $t_1^+ - t_2^+$ fluctuates between 0 and 20, with these values being randomly generated from trial to trial. This approach enables to identify the optimal parameters that yield superior performance across all potential delay scenarios. This performance assessment is designed to quantify the extent to which the models demonstrate temporal task generalization capabilities. 

\subsection{Analysis}
After optimization, the models are trained to select the position associated with best stimuli (equivalent to the most rewarding one). The training procedure consists of 1000 trials, with each trial being randomly chosen from the 72 stimulus-position pairs. The timings and delays for each stimulus are also randomly determined. Following the training phase, the models undergo testing on 1000 randomly selected trials. The overall performance is measured as the proportion of correct choices out of the 1000 trials. The results are further analyzed by separating two scenarios: when the best
stimulus appears first and when the best stimulus appears last. The first scenario enables an evaluation of the models' working memory: if the best stimulus emerges first, the model must retain its value until the end of the trial. The second scenario allows an evaluation of performance when the model needs to respond rapidly: if the best cue appears last, the model must quickly adjust its decision before the trial ends. This entire process is repeated across 10 different seeds for each model, and the final performance is determined as the average across the seeds. A paired t-test is employed to assess whether there is significant differences between the performances of the model $M^0$ and the other ones.

\section{Results}
The results depicted in Figure \ref{fig:results} indicate that the models $M^2$, $M^3$, and $M^*$ exhibit significantly better overall performances compared to models utilizing a single reservoir such as $M^{0}$ (blue bars). The paired t-test with $M^0$ results in p-values of 3e-4, 3e-5 and 1e-8 respectively.  These three models demonstrate overall better performances primarily because they outperform in the scenario where the best cue appears first (green bars), whereas no particular difference is observed in the scenario when the best cue appears last (yellow bar).  These models have the common thread of being composed of two pathways with distinct average leak rates that emerged from the hyperparameter optimization of the models, and depicted in Table \ref{table:leak_rates}. The difference in values of leak rates between pathways enables different speed of treatment: the latest cue is always fed into the pathway P2 that has a faster processing thanks to a bigger leak rate. Conversely, the earliest cue is always fed into the pathway P1 that has a slower processing thanks to a small leak rate.  Furthermore, the overall performances are improving as the depth of the pathways is increasing, until reaching the continuous limit with $M^*$. The latter achieves the best  performances with 89.5\% of success, outperforming the classical reservoir with 74.0\% of success. This improvement is also mostly visible in trials when the best cue appears first, going from 68.2\% of success with $M^1$ to 87.8\% of success with $M^*$. The difference in performances is visible during the training process as depicted in Figure \ref{fig:training}, where the dual models are learning faster as the depth of the paths is increasing. 

\begin{figure}[h!]
\includegraphics[width=0.95\columnwidth]{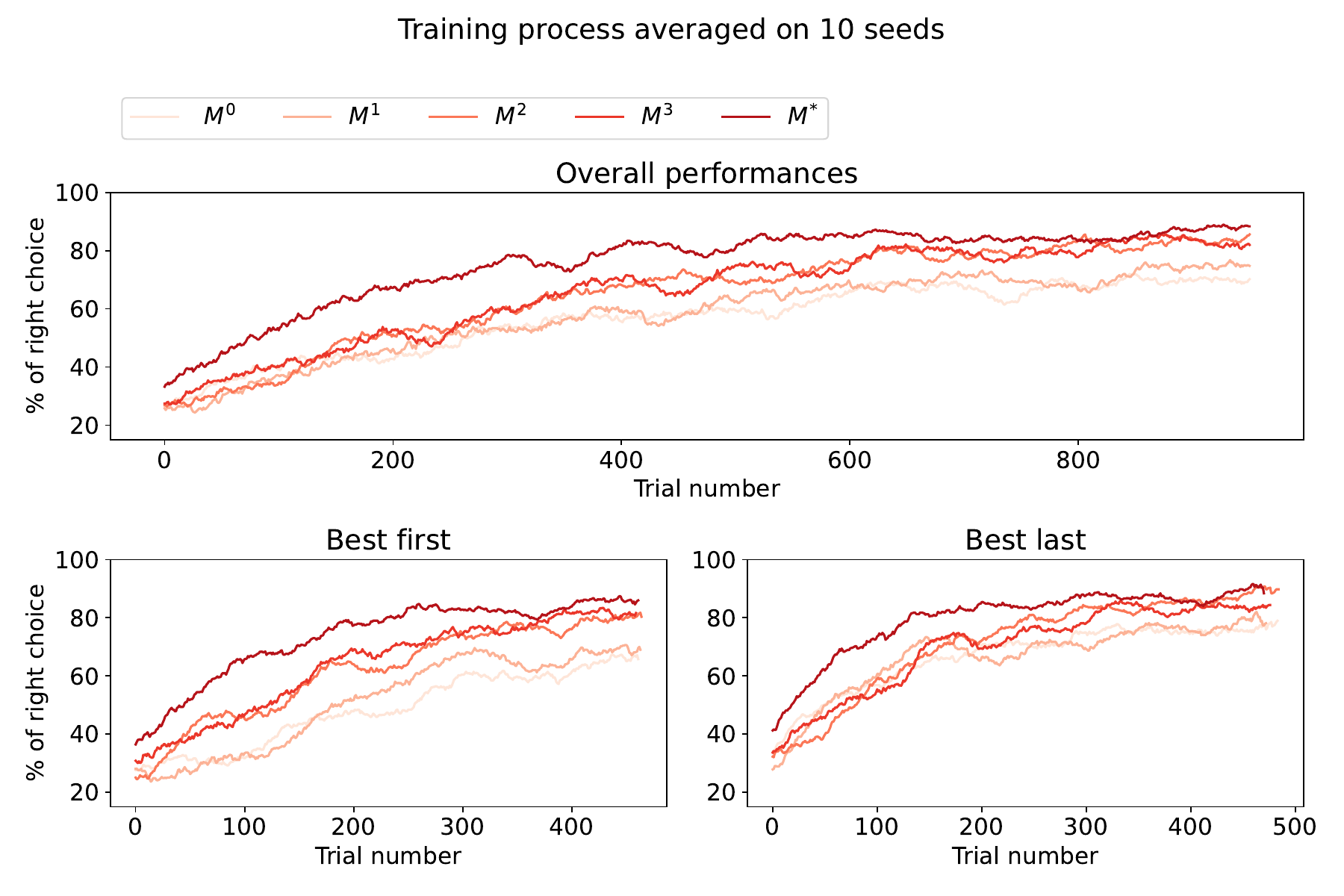}
\caption{\textbf{Top} Training process. The curves correspond to the percentage of successful choice using a moving average that takes the 50 last trials. \textbf{Bottom} The training process is categorized in two scenarios based on when the best cue appears. This enables to observe that the major difference in performances occurs when the best cue appears first.}
\label{fig:training}
\end{figure}

\begin{figure}[h!]
\includegraphics[width=\columnwidth]{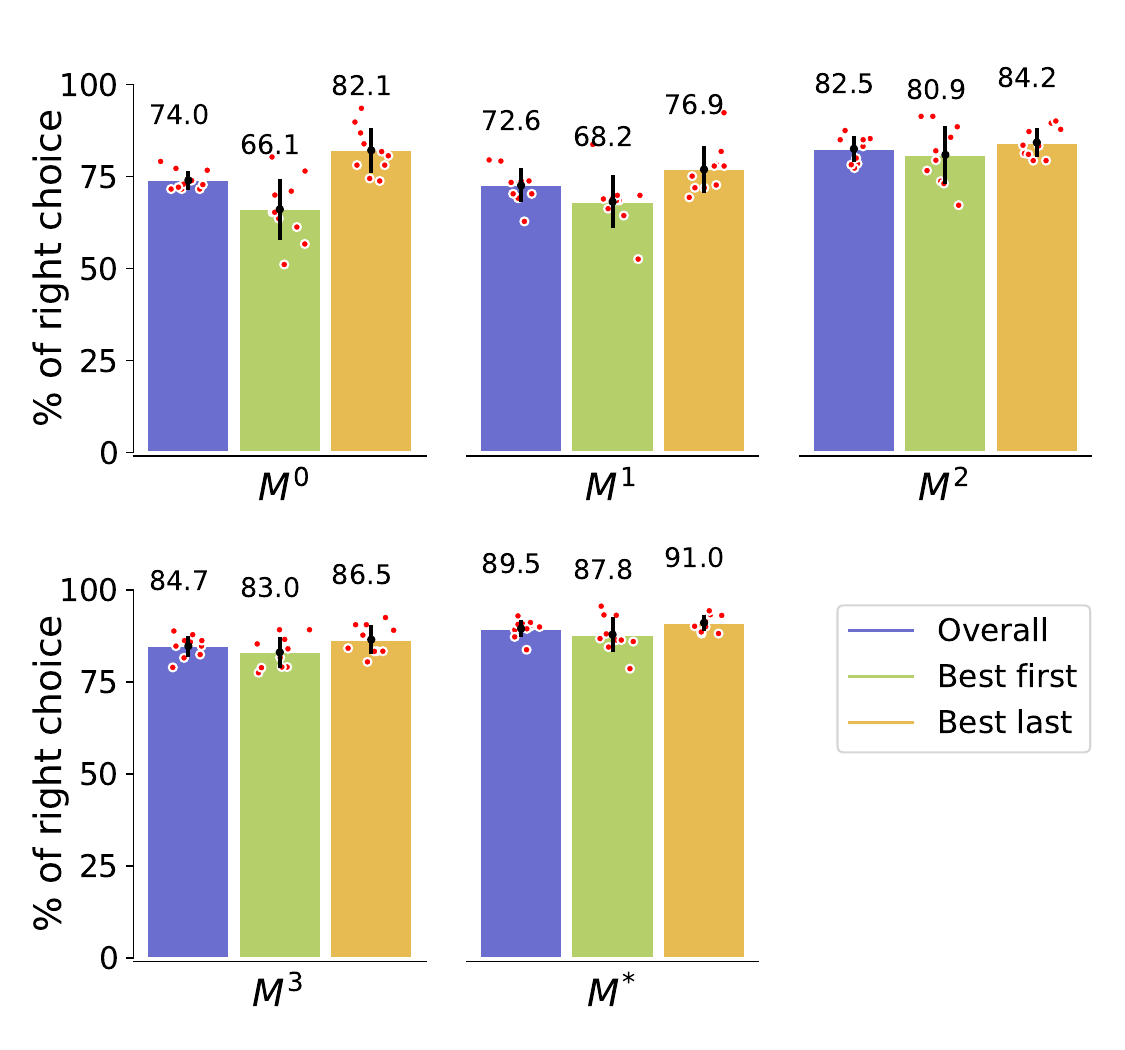}
\caption{Performance comparison of the models. The blue bar corresponds to the percentage of successful choice out of the 1000 tested trials. This result is categorized in two cases: when the best cue appears first (green) and when the best cue appears last (yellow). Models with dual pathway and sufficient depth such as $M^2$, $M^3$ and $M^*$  gives significantly better performances overall and when the best cue appears first. $M^*$  corresponds to the deepest dual pathway and demonstrates optimal performances.}
\label{fig:results}
\end{figure}

\begin{table}[htbp]
\centering
\begin{tabular}{l|l|l|l}  
\textbf{Model} &  \textbf{Pathway 1 (P1)} & \textbf{Pathway 2 (P2)} & \textbf{P2 / P1}\\ 
\hline
{\textbf M$^\mathbf{0}$} &  0.06 & --- & --- \\ 
\hline
{\textbf M$^\mathbf{1}$} &  0.068 & 0.67  & $\approx$ 9.8\\ 
\hline
{\textbf M$^\mathbf{2}$} &  0.17 (0.06, 0.28) & 0.29 (0.50, 0.07)  & $\approx$ 3.6 \\ 
\hline
{\textbf M$^\mathbf{3}$} & 0.23 (0.16, 0.10 ,0.43) & 0.59 (0.07, 0.72, 0.99)  & $\approx$ 2.5  \\ 
\hline
{\textbf M$^\mathbf{*}$} &  0.23 & 0.59  & $\approx$ 2.5 \\ 
\hline
\end{tabular}
\caption{Mean value of the leak rates ($\alpha$ of equation \ref{eq:train_1}) for pathway 1 \& 2 in the models. The ratio $P2/P1$ is highlighting the existence of a fast (characterized by a big leak rate) and a slow pathway (characterized by a low leak rate). }
\label{table:leak_rates} 
\end{table}

\section{Discussion}

Starting from a trivial and abstract non-stochastic two-arm bandit task, we complexified the task by introducing a motor indirection as well as a temporal component (embodiment). The initial task (no motor indirection, no temporal aspect) can be fully solved by an ESN in just a few trials using reinforcement learning (with a success rate of 96.7\% when motor indirection is absent and 99\% when temporal aspect is absent). However, as soon as motor indirection and temporal aspect are jointly introduced, performances dropped drastically, especially in the case when the best option is presented first. These surprising results can be understood when considering a few representatives cases (see also Fig.~\ref{fig:output}):
\begin{figure}[h!]
\includegraphics[width=\columnwidth]{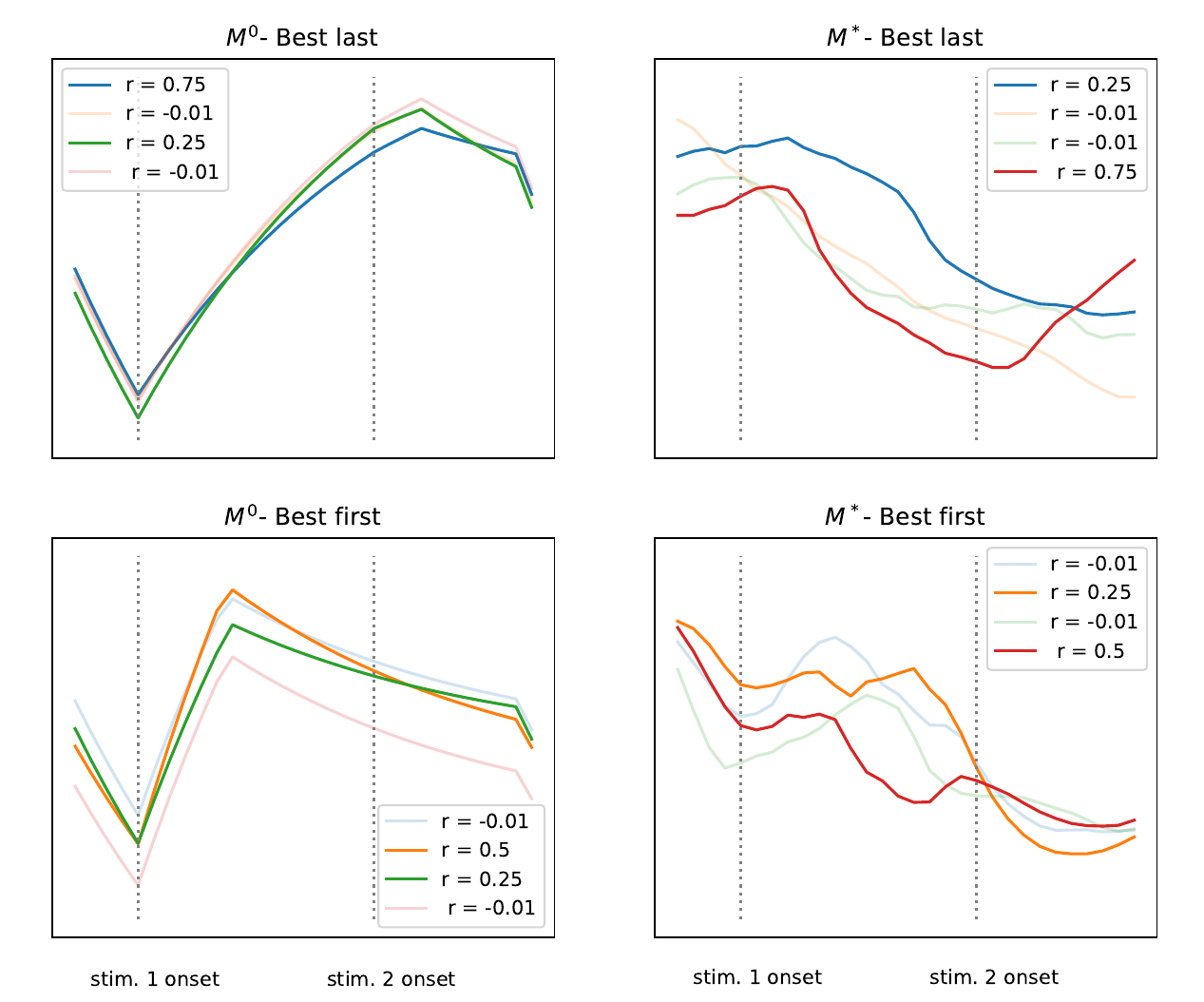}
\caption{Output activity during challenging trials where the reward have close values. \textbf{Top} The \textbf{M}$^\mathbf{0}$ fails at processing the best cue that arrives late, while \textbf{M}$^\mathbf{*}$ model is able to react quickly with the late best option and to choose the corresponding best motor action. \textbf{Bottom} The \textbf{M}$^\mathbf{0}$ fails at retaining the best cue that arrives first while \textbf{M}$^\mathbf{*}$ model successfully recalls the best option and to choose the corresponding best action.}
\label{fig:output}
\end{figure}

\begin{itemize}
    \item When the best option is presented first, it is not sufficient to memorize the motor action to be made because when the second stimuli arrives, the expected value of the first stimulus (as computed internally by the model) needs to be compared to the expected value of the second stimulus such as to make the right motor command.
    \item When the best options is presented late, the model has only a few time steps to process the new option in order to find the related expected value, to compare this value to the current one and finally, to make a motor decision that needs to overcome the alternative choice.
\end{itemize}
Said differently, the model needs to be reactive for some trials and conservative for some others. Results displayed on figure \ref{fig:results} clearly indicate that this is hardly the case for the regular M$^\mathbf{0}$ model, with a mean performance of 74\%. However, as soon as we introduce a dual pathway architecture, performances increases with the depth of the pathway, best performances being achieved by the  continuous M$^\mathbf{*}$.

Table \ref{table:leak_rates} displays the mean leak rates over the two pathways in all models (when relevant). Interestingly enough, the second pathway, that is, the one receiving the late stimulus has a much stronger leak rate when compared to the first pathway. This means units in the second pathway are able to process information much more rapidly when compared to the first pathway, even tough this also mean they have a reduced memory capacity (because they leak past activity at a high rate). Overall, all the dual pathways models (M$^\mathbf{1}$, M$^\mathbf{2}$, M$^\mathbf{3}$, M$^\mathbf{*}$) developed both a slow and fast pathway resulting from the optimization process. If we now turn back to our initial inspiration on the structure of the basal ganglia, this dual slow/fast pathway is reminiscent of the direct and hyperdirect pathway even though we do not pretend for equivalence.

%
%



\bibliography{references.bib}
\end{document}